\documentclass[runningheads]{llncs}
\usepackage{graphicx}
\usepackage{amsfonts}
\usepackage{amstext}
\usepackage{amsmath}
\usepackage{xcolor}
\usepackage{mathpazo}
\usepackage{amsmath}
\usepackage{wrapfig}
\usepackage{multirow}
\usepackage{graphics}
\usepackage{array}
\usepackage{adjustbox}

\begin{document}
	\title{InceptionGCN: Receptive Field Aware Graph Convolutional Network for Disease Prediction}
	\titlerunning{InceptionGCN }
	%
	\author{Anees Kazi\inst{1} \and
		Shayan Shekarforoush\inst{2} \and
		S.Arvind krishna\inst{3} \and
		Hendrik Burwinkel\inst{1} \and
		Gerome Vivar\inst{1,4} \and
		Karsten Kort\"um \inst{5} \and Seyed-Ahmad Ahmadi\inst{4} \and Shadi Albarqouni \inst{1} \and Nassir Navab \inst{1,6}}
	\authorrunning{A. Kazi et al.}
	%
	\institute{Computer Aided Medical Procedures (CAMP), Technical University of Munich, Munich, Germany \and Sharif University of Technology, Tehran, Iran \and Department of Computer Science and Engineering, National Institute of Technology Tiruchirappalli, India \and German Center for Vertigo and Balance Disorders, Ludwig Maximilians Universit\"at M\"unchen, Germany \and Augenklinik der Universit\"at, Klinikum der Universit\"at M\"unchen, Germany \and Whiting School of Engineering, Johns Hopkins University, Baltimore, USA
	}

	%
	\maketitle              
	\begin{abstract}
		Geometric deep learning provides a principled and versatile manner for integration of imaging and non-imaging modalities in the medical domain. 
		Graph Convolutional Networks (GCNs) in particular have been explored on a wide variety of problems such as disease prediction, segmentation, and matrix completion by leveraging large, multi-modal datasets. In this paper, we introduce a new spectral domain architecture for deep learning on graphs for disease prediction. The novelty lies in defining geometric 'inception modules' which are capable of capturing intra- and inter-graph structural heterogeneity during convolutions. 
		We design filters with different kernel sizes to build our architecture. We show our disease prediction results on two publicly available datasets. Further, we provide insights on the behaviour of regular GCNs and our proposed model under varying input scenarios on simulated data.
	\end{abstract}
\section{Introduction}
There is an increasing focus on applying deep learning on unstructured data in the medical domain, especially using Graph Convolutional Networks (GCNs) \cite{defferrard2016convolutional}. Multiple applications have been demonstrated so far, including Autism Spectrum Disorder prediction with manifold learning to distinguish between diseased and healthy brains \cite{ktena2018metric}, matrix completion to predict the missing values in medical data \cite{vivar2018multi}, and finding drug similarity using graph auto encoders \cite{ma2018drug}. In this paper, we study the task of Alzheimer and Autism disease prediction with complementary imaging and non-imaging multi-modal data.
	
In above works, GCNs had a remarkable impact on the usage of multi-modal medical data. One key difference to previous learning-based methods is to set patients in relation to each other with a neighborhood graph, often by associating them through non-imaging data like gender, age, clinical scores or other meta-information. On this graph, patients can be considered as nodes, patient similarities are represented as edge weights and features from e.g. imaging modalities are incorporated through graph signal processing. GCNs then provide a principled manner for learning optimal graph filters that minimize a objective. Here, we use node-level classification for our disease prediction task.
	
A simple analogy to node-based classification of the population is image segmentation with CNNs, where each pixel is a node and the image grid is the graph. In such domains, filters with a constant size can manage to acquire semantic features over the whole grid domain, given convolutions over a constant number of equidistant neighbors. In the case of irregular graphs, the number of neighbors and their distance from each other leads to heterogeneous density and local structure. Applying filters with constant kernel size over the whole grid domain might not produce semantic and comparable features. 
	
\begin{figure*}[t!]
		\centering
		\includegraphics[width=1.0\textwidth]{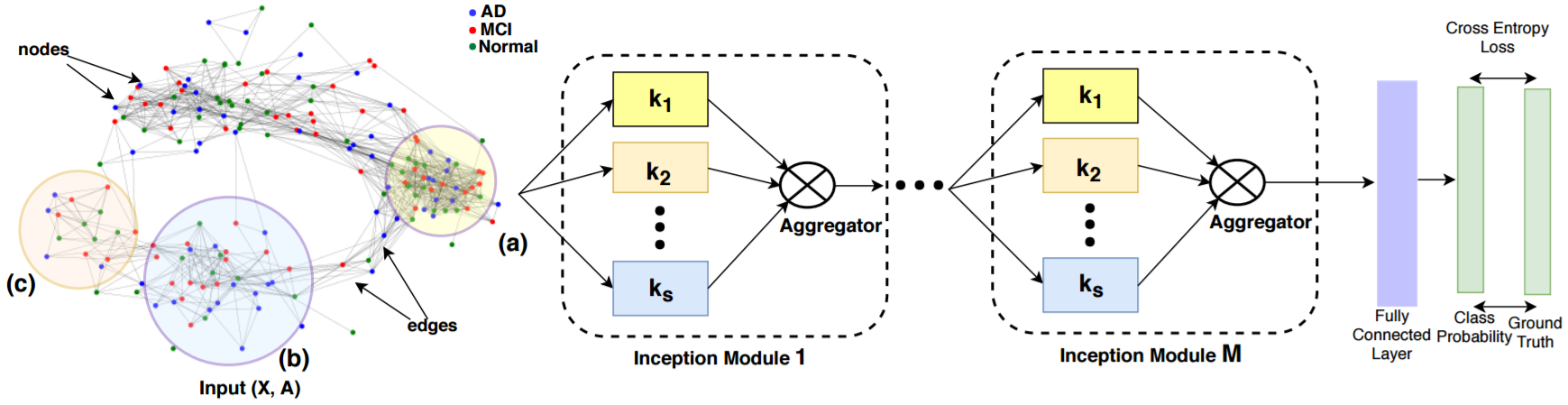}
		\caption{Left: Affinity graph with clusters for TADPOLE dataset, different cluster sizes are depicted at points (a), (b) and (c). Right: Setup of InceptionGCN, feature matrix $X$ is processed by several GC-layers with considered neighborhood $k_1, \cdots, k_S$ in each inception module. The output of each layer is used in the aggregator function.}
		\label{Inc_Net}
	\end{figure*}

In medical datasets, graphs defined on patient's data observe similar heterogeneity, as each patient may have a distinct combination of non-imaging data and different number of neighbors. A concrete example is shown in Fig. \ref{Inc_Net} (left), which depicts a population graph of 150 subjects for Alzheimer's disease classification, who are arranged in clusters of varying density and local topology (regions a, b and c). 
Such heterogeneity in the graph structure should be considered to learn cluster-specific features. A model capable of producing similar intra-cluster and different inter-cluster features can be designed by applying multi-sized kernels on the same input. To this end, we propose InceptionGCN, inspired by the successful inception \cite{szegedy2015going} architecture for CNNs. Our model leverages spectral convolutions with different kernel sizes and chooses optimal features to solve the classification problem.

To the best of our knowledge, there is not much related literature that focuses on receptive fields of GCN filters. Earlier works \cite{defferrard2016convolutional,kipf2016semi} use GCNs with constant filter size for the node-based classification task and show the superiority of GCN but do not address the problem of heterogeneity of the graph. In \cite{liu2018geniepath}, a method is proposed that determines a receptive path for each node rather than a field for performing the convolutions for representation learning. Irrespective of nearest neighbors, the aim is to perform convolutions with selective nodes in the receptive field. 
In \cite{xu2018representation}, a DenseNet-like architecture \cite{huang2017densely} is proposed, in which outputs from consecutive layers are concatenated. Here, the receptive field is addressed in an indirect way since the output features of successive layers depend on multiple previous layers through skip connections. Another work \cite{hamilton2017inductive} uses features that are either fixed, hand-designed or based on aggregator-functions. Moreover, the method needs a pre-defined order of nodes which is difficult to obtain. 
	
In this paper we show that InceptionGCN  is an improvement in terms of performance and convergence. Our contributions are: (1) we analyze the inter-dependence of graph structure and filter sizes on one artificial and two public medical datasets and in doing so, we motivate the need for multiple kernel size. (2) We propose our novel InceptionGCN  model with multiple filter kernel sizes. We validate it  on artificial and clinical data and the show improved performance over  regular GCN architectures. (3) We demonstrate the robustness of our model towards different approaches for constructing graph adjacency from non-imaging data. 
\vspace{-0.3cm}
\section{Methodology}
Traditional models \cite{parisot2017spectral} use a constant filter size throughout all layers, which forces the features of every node to be learned using neighbors at a fixed number of hops away without consideration of cluster size and shape. Our proposed InceptionGCN  model overcomes this limitation by varying the filters' size across the GC-layers in order to produce class separable output features. This property of our model is highly desirable when each class distribution has distinct variance and/or when the classes are heavily overlapping. Utilizing this setting, we target to solve the disease classification task by incorporating semantics of varied associations coming from different graphs within the population. We provide a detailed description of the model starting from the affinity graph construction followed by the mathematical background and a discussion of the proposed model architecture.
	\vspace{-0.2cm}
	\subsection{Affinity graph construction}  
The construction of an affinity graph is crucial to accurately model the interactions among the patients and should be designed carefully. 
The affinity graph $G = \left ( V,E,W \right )$ is constructed on the entire population (including training and testing samples) of the patients, where $\left | V \right | $= N vertices, $E$ are the edge connections of the graph and $W$ are the weights of the edges. Considering each patient as a node $n_{i}$ in the graph, $G$ incorporates the similarities between the patients with respect to the non-imaging data $\eta$. 
The features $x_{i} \in \mathbb{R}^{N} $ at every node $n_i$ are fetched from imaging data. First, we construct a binarized edge graph $E \in \mathbb{R}^{N\times N}$ representing the connections. Mathematically, $E$ can be defined as 
	\begin{equation}\label{eq:1}
	\begin{aligned}
	E_{i,j} = \left\{\begin{matrix}
	1 & if \left | \eta _{i} - \eta _{j}\right |< \beta \\ 
	0 & otherwise
	\end{matrix}\right. 
	\end{aligned}
	\end{equation}
where  $\eta_{i}$ and $\eta_{j}$ are the values of the non-imaging element for nodes $i$ and $j$ and $\beta$ is the threshold for that element. The weight matrix $W \in \mathbb{R}^{N\times N}$ weights the edges based on the correlation distance between the features at every node. The weight matrix elements are defined as $W_{i,j} = Sim\left ( x_{i},x_{j} \right )$ ,where $Sim(x_{i}, x_{j}) = exp(-\frac{\left [ \rho(x_{i},x_{j}) \right ]^{2}}{2\sigma ^{2}})$ with $\rho$ being the 'correlation distance' and $\sigma$ being the width of the kernel. This weight computation and value of $\beta$ is identical to the procedure described in \cite{parisot2017spectral}, to provide equal grounds for comparison.  
The final affinity matrix $A$ is constructed as $A = W\circ E$ with $\circ$ being the Hadamard product.
\subsection{Mathematical background of spectral convolution and localization of filters for inception modules}
Let $L = I_{N}- D^{-\frac{1}{2}} A D^{-\frac{1}{2}}$ be the normalized version of the graph Laplacian of $G$ including self loops. $D$ is the diagonal matrix with $D_{ii}=\sum_{j}A_{ij}$, $I_{N} \in \mathbb{R}^{N\times N}$ being the identity matrix.  Since $L$ is real positive and semi-definite, it is diagonalizable by its eigen vectors $U \in \mathbb{R}^{N\times N}$ such that $L = U\Lambda U^{T}$, where $\Lambda = diag(\lambda _{0},\lambda _{1},...\lambda _{N-1}) \in  \mathbb{R}^{N\times N}$ are the corresponding eigen values. The graph Fourier Transform of a signal $x $ at each node is defined as $\widehat{x} = U^{T}x \in \mathbb{R}^{N}$, the inverse Fourier Transform as $x= U\widehat{x} \in \mathbb{R}^{N}$. With this information, the spectral convolution can be defined as a multiplication of the signal $x$ with a learnable filter $g_{\theta }= diag(\theta)$ in the Fourier domain, which results in $y = U g_{\theta }\left ( \Lambda  \right )U^{T}x = g_{\theta }\left ( U\Lambda U^{T} \right )x = g_{\theta }\left ( L \right )x$ interpreting $g_{\theta}$ as a function of the eigenvalues $\Lambda$ \cite{kipf2016semi}. In order to prevent the computationally prohibitive matrix multiplication necessary to perform the Fourier Transform of signal $x$, we redefine $g_{\theta}$ using the Chebyshev polynomial parameterization of the filter $g_{\theta }\left ( \Lambda  \right ) = \sum_{r=0}^{k}\theta _{r} T_r(\Lambda)$, where $\theta \in \mathbb{R}^{k} $ is a vector of Chebyshev coefficients with degree $k$ \cite{kipf2016semi,defferrard2016convolutional}. 
Since $L^k = (U\Lambda U^{T})^k = U\Lambda ^{k}U^{T}$, we can write $g_{\theta }\left ( \Lambda  \right )$ as a function of $g_{\theta }\left ( L  \right )$. Therefore, we can perform the spectral filtering on a signal $x$ with $g_{\theta} \ast x = \sum_{r=0}^{k}\theta _{r} T_r(L) x$. The value of vertex $j$ of the filter $g_{\theta}$ centered at vertex $i$ is given by 
\begin{equation}
(g_{\theta }(L)\delta_{i})_{j} = (g_{\theta }(L))_{ij}\\
= \sum_{k}\theta _{k}(L^{k})_{ij}
\end{equation}
where $\delta_{i}$ is Kronecker delta function. Inspired by \cite{hammond2011wavelets}, here we explain how the filters of specific receptive fields can be derived. Let $G$ be a weighted graph,  $L$ be the graph Laplacian (normalized or unnormalized), and $k>0$ be an integer (here $k$ stands for the $k^{th}$ hop neighbor), then for any two vertices $i$ and $j$:
\begin{equation}
\left ( L^{k} \right )_{ij}  = \left\{\begin{matrix}
\Omega & \: d_{G}(i,j) \leq k \\
0& \: otherwise
\end{matrix}\right.
\end{equation}
where $d_{G}(i,j)$ is the shortest path distance between $x_{i}$ and $x_{j}$ and $\Omega$ is the sum of all edge weights on the shortest path from $x_{i}$ to $x_{j}$.
Therefore from eq. 2 the spectral filters represented by $k^{th}$ order polynomial of the Laplacian are exactly $k$-hop localized.
\vspace{-0.4cm}
\subsection{Inception modules}
The localization of a filter is defined by taking  all the neighbors at a distance of $k$ hops into account for the spectral convolution with a signal x. A filter $s$ with a fixed $k_s$ used on the full dataset $X$ can be defined as $y_s =  \sum_{r=1}^{k_s} T_{r}(L) X \theta_{r,s}$. Here, $y_s$ describes the output of a filter with neighborhood in $k_s$-hop distance. To account for different sizes and variances of clusters and structure in the data, instead of using one filter we now use $S$ filters with varying neighborhood $k_{s}$. These combined filters $s$ are the centerpiece of the inception module as they simultaneously consider the close proximity of a signal $x$ and the broader neighborhood situation. Every filter of the module has its own parameter vector $\theta_s$ and performs a convolution on the dataset $X$ for returning an output vector $y_s$. The outputs of each filter are merged in an aggregator-function $\Psi$ to determine the output $y$ of the inception module as $y = \Psi \left (  y_1, \cdots, y_S \right )$
where every $\theta_s \in \mathbb{R}^{k_s}$ with entries $\theta_{r,s}$ is the learnable parameter vector for each filter of the inception module. To merge the output of each inception module we propose two aggregators $\Psi$, (1) concatenation and (2) max-pooling. Our model architecture is illustrated in Fig. \ref{Inc_Net}. It is built with $M$ inception modules. Each inception module consist of $S_m$ GC-layers in parallel with filters of different $k_{s,m}$. We apply ReLU at the output of each GC-layer. For the training set, a labelled subset of graph nodes is chosen, for which the loss is computed and gradients are backpropagated. We apply cross-entropy loss as the optimization function. Due to the graph connections, the training process on the labelled data is transferred to the unlabeled data by signal diffusion which corresponds to the behavior of a standard GCN.
\section{Experiments and Results } 
In this section, we provide two main experimental setups to show (1) the sensitivity of spectral convolutions to different graphs and kernel sizes of the filters and (2) superiority of the InceptionGCN  to other baseline methods. We show our results on two multi-modal medical datasets and thoroughly analyze both the baseline \cite{parisot2017spectral} and the proposed model. At last, we provide insights into generalized design choices for building a data and task-specific model.
\subsection{Datasets}
\textbf{TADPOLE \cite{marinescu2018tadpole}:} This dataset is a subset of the Alzheimer's Disease Neuroimaging Initiative (adni.loni.usc.edu), consisting of 557 patients with 354 multi-modal features per patient. The target is to classify each patient into one of the three classes (Cognitively Normal (CN), Mild Cognitive Impairments (MCI) or Alzhe-imer's Disease (AD). Features are extracted from MR and PET imaging, cognitive tests, CSF and clinical assessments. The protein class APOE constitutes another factor assisting in patient classification. Testing this gene status provides a risk factor of developing AD. FDG-PET imaging measures the brain cell metabolism, where cells affected by AD show reduced metabolism. Furthermore, demographics are provided (age, gender). We construct a binarized graph with each element of demographic data, APOE status and FDG PET measures. We choose $\beta$ = 2 for age and $\beta$ = 0 for the rest of the three respectively. The edges are based on the $Sim(x_{i},x_{j})$ i.e. the feature similarity measure. We construct the 'Mixed' affinity graph by averaging all the graphs weighted with W and 'Mixed (no$Sim$)' without weighting.\\ 
\textbf{ABIDE \cite{abraham2017deriving}:} 
The Autism Brain Imaging Data Exchange (ABIDE) aggregates data from 20 different sites and openly shares 1112 existing resting-state functional magnetic resonance imaging (R-fMRI) datasets with corresponding phenotypic elements (gender) for 2 classes normal and with Autism Spectrum Disorder (ASD). We choose 871 subjects divided into normal(468) and ASD diseased (403) subjects. For fair comparison, we follow the same pre-processing step as performed in the baseline method \cite{parisot2017spectral}. 
We construct two affinity graphs for non-imaging elements, gender and site, by choosing $\beta$ = 0 for both graphs. 
\subsection{Experiments on medical datasets}
In this subsection we present both the experimental setups mentioned above and discuss our findings on the medical datasets.\\
\textbf{Effect of different kernel size on spectral convolution:}
Our first set of experiments is designed to investigate the optimal kernel size of the filter required for each graph. The baseline model \cite{parisot2017spectral} with two GC-layers in sequence is used to find out the required graph specific filter sizes (i.e. value of $k$). We investigate the performance of the model with the same input (features and graph) and $k_{1}$ and $k_{2} \in [1,6]$. Here $k$=1 and $k$=6 indicate the kernel size of one-hop (smallest) and six-hop neighbors (largest) respectively. We select the value of two $k$ corresponding to the best performance in the heatmap and incorporate them to our proposed InceptionGCN model as different kernel sizes. Like this, it is guaranteed that the sequential GCN is performing at its optimum when compared to our method. We discuss the validity of this setting in the later section.\\
\textbf{Results:}
	\begin{figure*}[t!]
		\begin{center}
			\includegraphics[width=1.0\linewidth]{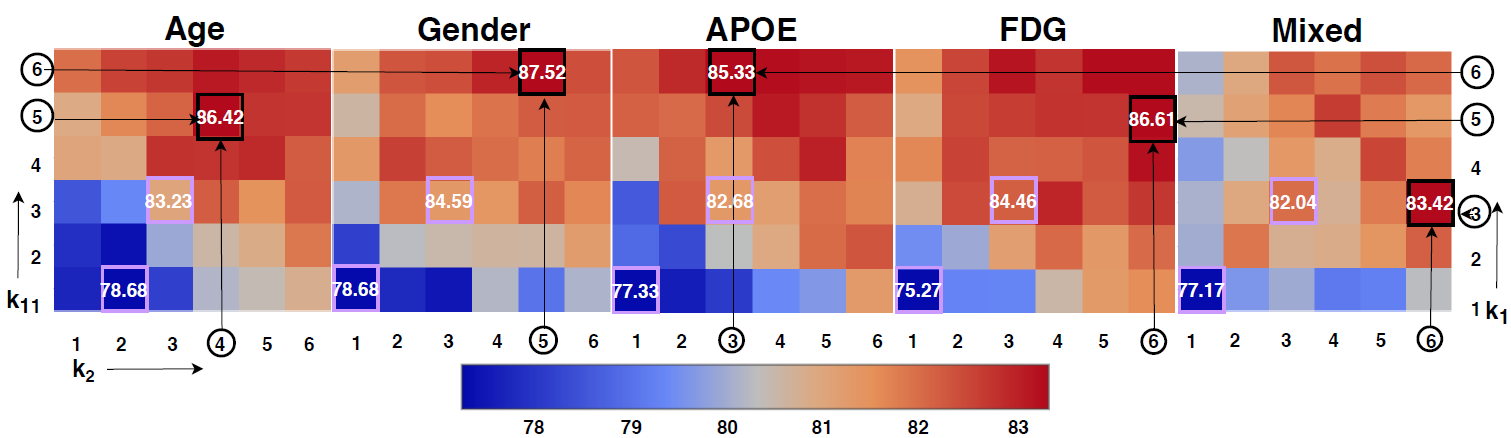}	\end{center}
		\caption{Heat maps for representing the performance of GCN on the TADPOLE dataset with varying kernel size of the filters. Each heat map comes from the distinct graph mentioned above. The highest and lowest performing combination of $k_{1}$ and $k_{2}$ are highlighted with a black box and the corresponding $k$-values are shown.}
		\label{fig:k1_k2}
	\end{figure*}
	Fig. \ref{fig:k1_k2} shows the corresponding results in terms of heatmaps.
Smaller $k$ learn local features and larger $k$ learn global features. The performance differs with the change in $k_{1}$ and $k_{2}$ by a margin of 8\% on average. It indicates that spectral convolution models are sensitive to the selection of $k$. The accuracy increases with the value of $k$, but becomes consistent with further increase. For most of the graphs $k_{1} > k_{2}$ is the best combination, since the initial layer filters look at global features.
Each affinity graph shows a different structure over the same vertices and shows varied results over the same combination of the two $k$. A similar trend is seen for ABIDE, which reassures the concept of sensitivity towards $k$.\\
\textbf{Comparison of InceptionGCN  against sequential GCN approaches:} We show the comparison with four baselines. Parisot et al. \cite{parisot2017spectral} is the traditional GCN with $k_{1}= k_{2}= 3$. We modify the same architecture of \cite{parisot2017spectral} with the best combination of the two $k$ mentioned as baseline $[k_{1},k_{2}]$. We evaluate our aggregator-function $\Psi$ for a proper selection of activations from all the individual GC-layers of the inception module by comparing them to the baseline $[k_{1},k_{1}]$ and $[k_{2},k_{2}]$. This comparison shows that $\Psi$ is not biased towards any particular kernel size. With such setting for all methods, each graph yields a different performance, showing the effect of the different neighborhood affinity as shown in Tab. \ref{tab:results_tadpole}. 
Our model outperforms the baselines \cite{parisot2017spectral} by an average margin of 4.12 \% for TADPOLE dataset.

The comparative results for ABIDE are given in Tab. \ref{tab:abide_comparative}. Our model performs comparable to the baseline \cite{parisot2017spectral}, but is not able to outperform it. Interestingly, the mixed graph with feature-based edge weighting performs worse than the weighting case. This confirms the non-discriminative nature of the features. Images collected from different sites make it harder for the model to learn class-discriminative features.
\vspace{-0.4cm}
\subsection{Experiments on simulated data} 
Seeing the contradictory performance on the two datasets, we investigate the model in detail for better understanding of the spectral model and to interpret better design choices for user-specific tasks. These experiments are specifically designed to investigate only the choice of the kernel size of the filters.
	
\begin{figure*}[t!]
		\begin{center}
			\includegraphics[width=1.0\linewidth]{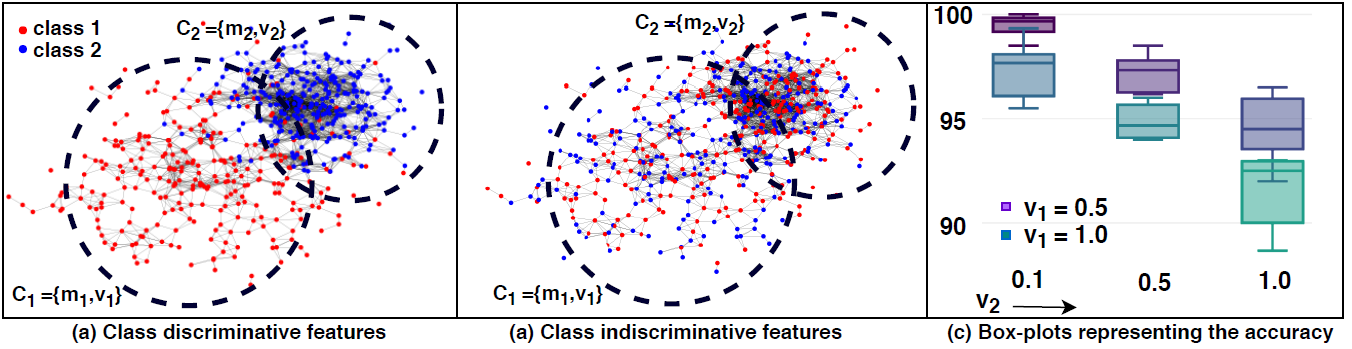}	\end{center}
		\caption{(a) represents the scenario of simulated data, where we change variances $v_{1}$ and $v_{2}$, (b) shows the scenario where the features are sampled from random distribution, (c) shows the variation in the performance in terms of accuracy for all the combinations of $v_{1}$ and $v_{2}$ for scenario (a).}
		\label{simulated_data}
	\end{figure*}

We generate two 2-dimensional clusters $C_{1}$ and $C_{2}$ having normal Gaussian distributions with 300 points each in Euclidean domain, each distribution representing one class. We construct the graph based on Euclidean distance between the features and $\beta = 0.5$ to sparsify the graph. This represents that the graph is highly correlated to the labels. In order to keep the experiment easy to interpret, we set means $[m_{1},m_{2}]$=$[-1,1]$ for $C_{1}$ and $C_{2}$ respectively and vary the corresponding variances $v_{1}$ and $v_{2}$. 
For features we show two settings: \textbf{class-discriminative}, where the (x,y)-values of the location of each point are considered as features and \textbf{class-indiscriminative}, where we randomly sample the features from a uniform distribution for both classes. Both settings are shown in Fig. \ref{simulated_data} (a) and (b). For the model architecture, we keep $M$ = 1 for both the baseline model \cite{parisot2017spectral} and InceptionGCN  and train both the networks at 200 epochs, with learning rate=0.2.
	\\
\textbf{Results and interpretation:}
The results of this experiment are illustrated with boxplots in Fig.\ref{simulated_data} (c). Each box shows the accuracy of the classification for different values of $k$ ranging from 1 to 10 for the baseline model for class-discriminative features. Keeping $v_{1}$ = 0.5, we vary $v_{2}$ for [0.1, 0.5, 1.0]. We repeat the experiments with $v_{1}$=1.0. It can be interpreted that when two clusters are clearly separable, the model is less sensitive to the value of $k$. Also it can be seen from the last two boxplots that with higher variance, the model becomes sensitive to $k$. Similar trends are observed when the value of $v_{1}$ is changed to 1.0, however a consistent drop in accuracy is observed with $v_{1}$ = 1.0. If there is large variance in the data, filters with larger receptive field will produce generalized global features.\\ 
Further, we apply our model to the simulated data with only one Inception module incorporating two GC-layers with different $[k_{1}, k_{2}]$=[1,10]. We compare the results of a single-layered GCN with $k$=[1,5,10] with the one layered inception module for four different settings. The superiority of our model is seen mainly in the challenging scenarios, where the variance of both classes is quite high (i.e. $v_{1}$ = $1.0$ and $v_{2}$ = $1.0$, cf. Tab. \ref{tab:good_bad_feat}). Here, we report the results for class indiscriminative features, where the performance drastically drops when features are totally random for all the models. InceptionGCN  outperforms the baseline in most of the cases. 
	\begin{center}
		\begin{table}[t]
			\caption{The performance of the model in terms of accuracy is represented in the table. $v_{1}$ and $v_{2}$ represent the variances of 2 classes of the simulated 2D Gaussian data. (a) In these cases the graph and corresponding features are highly correlated to the classes,  whereas in (b) only the graph is correlated to the classes.}
			\resizebox{\columnwidth}{!}{%
				\begin{tabular}{|c|p{3cm}|c|c|c|c|   }
					\hline
					\multicolumn{4}{|c|}{$v_{1}=0.5$}&\multicolumn{2}{|c|}{$v_{1}=1.0$}\\
					\hline
					&k &$v_{2}=0.1$&$v_{2}=1.0$&$v_{2}$=0.1&$v_{2}$=1.0\\
					\hline
					\multirow{ 5}{*}{}(a)&1 & \textbf{98.50 $\pm$ 01.38}& 94.50 $\pm$ 01.83  &\textbf{95.67 $\pm$ 02.49} & 92.50 $\pm$ 02.61 \\
					&10 & \textbf{99.00 $\pm$ 01.11}&93.67 $\pm$ 04.93&\textbf{95.50 $\pm$ 07.98}&91.00 $\pm$ 04.42 \\
					\hline
					&Inception-GCN (1 layer [k1,k2]=[1,10])  &94.83 $\pm$ 03.02&\textbf{97.00  $\pm$ 02.56}&92.00 $\pm$ 03.56&\textbf{94.33 $\pm$ 03.56 }\\
					\hline
					\hline
					(b)&1 &49.33 $\pm$ 06.84&50.33 $\pm$ 07.48&49.50 $\pm$ 04.60&50.00 $\pm$ 06.28 \\
					&10 &60.33 $\pm$ 16.78&53.50 $\pm$ 10.99&\textbf{50.83 $\pm$  06.02}&55.33 $\pm$ 14.79 \\
					\hline
					&Inception-GCN (1 layer [k1,k2]=[1,10])  &\textbf{66.50$\pm$ 17.12}&\textbf{64.00 $\pm$ 17.95}&48.00 $\pm$ 07.88&\textbf{69.00 $\pm$ 24.79}\\
					\hline
					\hline
			\end{tabular}}
			
			\label{tab:good_bad_feat}
		\end{table}
	\end{center}
	\begin{center}
		\begin{table}[t]
			\caption{Depicts the mean accuracies from stratified k-fold cross validation for all the setups of experiments for TADPOLE. The values of the chosen [$k_{1},k_{2}$] for the graphs are highlighted in the Fig. \ref{fig:k1_k2}.}
			\resizebox{\columnwidth}{!}{%
				\begin{tabular}{ |c|c|c|c|c|c|c|}
					\hline
					Affinity  &Age&Gender &APOE&FDG & Mixed&Mixed (no$Sim$)\\
					\hline
					\hline
					Parisot et al. \cite{parisot2017spectral} &82.55 $\pm$ 04.78&	84.59 $\pm$ 04.82&	82.68 $\pm$ 05.70	&84.46 $\pm$ 0 5.46&\textcolor{blue}{	82.04 $\pm$ 05.71}&82.11 $\pm$   04.94\\
					\hline
					\multicolumn{1}{|l|}{Baselines} & & &&&&\\
					$[k_{1},k_{2}]$ &86.42 $\pm$  03.95&	87.52 $\pm$ 03.51&	85.33 $\pm$ 04.75&	86.61 $\pm$ 04.53&	\textcolor{blue}{83.42 $\pm$ 05.93}&81.95 $\pm$ 05.92\\
					$[k_{1},k_{1}]$    &85.46 $\pm$ 05.6&86.19 $\pm$ 04.91&85.08 $\pm$ 05.21&86.55 $\pm$ 04.55&\textcolor{blue}{81.85 $\pm$ 06.28}&81.36 $\pm$ 05.98\\
					
					$[k_{2},k_{2}]$    &86.42 $\pm$ 03.98&84.59 $\pm$ 04.82&78.75 $\pm$ 04.45&84.46 $\pm$ 05.46&\textcolor{blue}{80.86 $\pm$ 05.69}&80.99 $\pm$ 04.71\\
					\hline
					\multicolumn{1}{|l|}{InceptionGCN } & & &&&&\\
					concat  &\textbf{88.35 $\pm$ 03.03}&	\textbf{88.06 $\pm$ 04.39}&	\textbf{88.14 $\pm$ 03.20}&	\textbf{86.99 $\pm$ 03.98}&	\textbf{\textcolor{blue}{84.35 $\pm$ 06.97}}&\textbf{83.62 $\pm$ 06.09}\\
					max-pool   &\textbf{88.53 $\pm$ 03.27}&	\textbf{88.19 $\pm$ 03.83}&	\textbf{88.49 $\pm$ 03.05}&	\textbf{87.65 $\pm$ 05.11}&	\textbf{\textcolor{blue}{84.11 $\pm$ 04.50}}&\textbf{83.87 $\pm$ 05.07}\\
					\hline
			\end{tabular}}
			\label{tab:results_tadpole}
		\end{table}
	\end{center}
	\begin{table}[b]
		\centering
		\caption{Depicts the mean accuracy from stratified k-fold Cross Validation for all the setups of experiments for ABIDE. The baseline values of [$k_{1}, k_{2}$] are [4,5], [6,5] and [4,4] for Gender, Site and Mixed, Mixed(no$Sim$) respectively.}
		\begin{tabular}{|c|c|c|c|c|}
			\hline
			Affinity  &Gender&Site& Mixed& Mixed(no$Sim$)\\
			\hline
			\hline
			Parisot et al \cite{parisot2017spectral}    &67.39 $\pm$ 04.76&67.39 $\pm$ 01.49&67.85 $\pm$ 00.63&\textcolor{blue}{69.80 $\pm$ 04.35}\\
			\hline
			\multicolumn{1}{|l|}{Baselines} & & &&\\
			$[k_{1},k_{2}]$    &\textbf{68.19 $\pm$ 05.38}&\textbf{69.00 $\pm$ 04.07}&\textbf{70.26 $\pm$ 03.70 }&\textbf{\textcolor{blue}{70.26 $\pm$ 04.58}} \\
			$[k_{1},k_{1}]$    &66.70 $\pm$ 06.90&68.65 $\pm$ 04.31& 69.91  $\pm$ 07.50&\textcolor{blue}{ 69.80 $\pm$ 03.90}\\
			$[k_{2},k_{2}]$    &65.78 $\pm$ 06.50&68.65 $\pm$ 04.31&69.00 $\pm$ 03.80& \textcolor{blue}{69.46 $\pm$ 04.69} \\
			\hline
			\multicolumn{1}{|l|}{Inception-GCN} & & &&\\
			concat   & 66.36 $\pm$ 05.66&67.97 $\pm$ 04.43&66.70 $\pm$ 06.27 &\textcolor{blue}{69.23 $\pm$ 06.66}\\
			max-pool &67.05 $\pm$ 05.47&67.39 $\pm$ 05.80&66.02 $\pm$   05.92&\textcolor{blue}{69.11 $\pm$ 06.68} \\
			\hline
		\end{tabular}
		
		\label{tab:abide_comparative}
	\end{table}
\vspace{-2.0cm}
\section{Discussion and Conclusion}
In this work we have introduced InceptionGCN, a novel architecture that captures the local and global context of heterogeneous graph structures with multiple kernel sizes. 
The validation included an investigation of spectral convolution parameters and the behaviour of the proposed model given varying input data, in comparison to a recently proposed baseline method \cite{parisot2017spectral}.  
	\begin{figure*}[t]
		\begin{center}
			\includegraphics[width=0.8\linewidth]{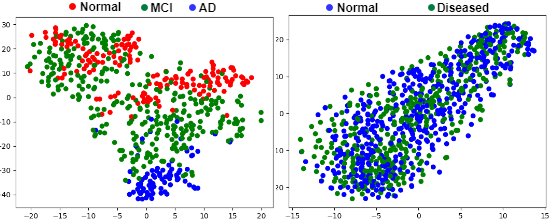}	\end{center}
		\caption{TSNE embedding in 2-dimensional space visualized on raw features for TADPOLE(left) and ABIDE(right) datasets. }
		\label{fig:embedding_input}
	\end{figure*}
Our findings show that applying different sized filters on the same input features and graph improves the process of feature learning at multi-scale levels. Such rich and heterogeneous features help the model to learn better filters for classification. We tested the method on two publicly available medical datasets for Alzheimer's and Autism disease prediction, in order to analyze the robustness of the model towards different features, graph affinities and tasks. Our results show that both the spectral convolution and the proposed model obtained high classification accuracies for TADPOLE (cf. Tab. \ref{tab:results_tadpole}), with a clear margin of InceptionGCN  over the baselines. In the case of the ABIDE dataset, however, both methods had comparable performance, which was considerably lower than on TADPOLE (cf. Tab. \ref{tab:abide_comparative}). To investigate the different performances of both models, we utilized simulated data with i) different degrees of class overlap in the feature space and ii) entirely random features, forcing the GCN models to rely on connectivity alone (Tab. \ref{tab:good_bad_feat}). It can be concluded that while both GCN models are very sensitive to variance of data,  our model shows the superiority in case of having large variances and overlapping of class clusters. The main factors affecting the performance of GCN are features, graph and filters. With all the experiments we discuss all the factors in detail.\\ 
\textbf{Influence of the graph:} 
For the ABIDE dataset, images are collected from 20 different sites and imaging conditions, which adds considerable heterogeneity to the data. Consequently, the affinity graph based on site information consists of 20 disjoint clusters. Building a graph based on site information allows only the neighbors (i.e. samples from the same site) to contribute to the feature learning. This has less clinical relevance to the classification task, whereas for TADPOLE, the risk factors and demographics are clinically relevant. Such relevance of the graph can be determined using the graphs' energy function provided in \cite{gansner2013coast}.
Next, the mixed affinity graph performs worst overall in terms of accuracy (cf. Tab. \ref{tab:results_tadpole} and Tab. \ref{tab:abide_comparative}) and Standard Deviation (SD) (cf. Tab. \ref{tab:abide_comparative}). This indicates that a straightforward creation of the mixed affinity graph by averaging impairs the inherent structure of each graph, and important clinical semantics from individual graphs may get lost. This is confirmed by the unequal performance observed for each affinity graph, which may even indicate a ranking of relevance of each non-imaging element to the objective. A more elegant way to combine all the affinity graphs is by ranking them while training \cite{kazi2018multi}.\\
\textbf{Influence of the features:}
The importance of a proper feature choice becomes clear in the tests on simulated data. When using randomly sampled features for every node (cf. Tab. \ref{tab:good_bad_feat}) the overall performance drops drastically. A large standard deviation in the performance shows that filters are not learned properly and the model does not converge. The same behavior can be seen for the TADPOLE and ABIDE dataset when comparing the mixed and mixed (no$Sim$) (cf. Tab.  \ref{tab:results_tadpole} and \ref{tab:abide_comparative}). Since the features of the ABIDE dataset are not distinguishing the nodes into different clusters compared to the TADPOLE dataset (Fig. \ref{fig:embedding_input}), the performance of the models drops for ABIDE when using the feature similarity ($Sim$), which is used for graph construction. At the same time, the models receive a performance boost when the meaningful features of TADPOLE are included into the graph generation process.\\ 
\textbf{Influence of the kernel size:}
We investigated the effect of features and heterogeneity of the graph towards the choice of $k$. Our results show that in case of class separable features, a larger value of $k$ will give more compact features. From Tab. 3, it is clear that InceptionGCN performs better in case that the classes have large and different variances. In such a case, InceptionGCN  with multiple $k_{s}$ manages to capture the class discriminative features for the nodes. If the clusters are compact (v=0.1) the choice of $k$ does not matter. From Fig. \ref{simulated_data} (c), we see that the model is not sensitive to $k$ if the clusters are compact, whereas it becomes sensitive when the variance increases. In case of class indiscriminative features and a less relevant graph (as is the case of ABIDE) a larger kernel size helps to learn global class discriminative feature.\\
\textbf{Sequential model vs. InceptionGCN:}
Choosing the values of the two $k$ from sequential model (GCN) for a parallel setting might seem ambiguous. In Tab. \ref{tab:results_tadpole},
the role of the aggregator-function is clearly visible in the performance, since the baselines are all the possible combinations that the final output of our model can get. Furthermore, our proposed model converges 1.63 times faster in terms of epochs compared to the baseline method when trained with early stopping criteria with window size of 25 due to a better feature learning process. \\
\textbf{Future scope:} Potential improvements of the InceptionGCN  model include out-of-sample inference (i.e. inductive learning), which will highly improve the usability of the model. Another area of investigation is the integration of multiple affinity graphs into one model. Furthermore, the InceptionGCN  model structure itself can also be optimized, first by using a learnable pre-processing step to obtain the neighborhood values $k$, and second, by analyzing the number of hidden units in each GC-layer and the overall number of inception modules necessary. 
\section{Acknowledgement}
The authors would like to thank Dr. Benedikt Westler for his help and support in understanding the TADPOLE dataset. The study was carried out with financial support of Freunde und F{\"o}rderer der Augenklinik, M{\"u}nchen, Germany, Carl Zeiss Meditec AG, Oberkochen, Germany and the German Federal Ministry of Education and Research (BMBF) in connection with the foundation of the German Center for Vertigo and Balance Disorders (DSGZ) (grant number 01 EO 0901).	
\vspace{-0.2cm}
\bibliographystyle{ieeetr}
\bibliography{biblio}
	
\end{document}